# Few-Shot Hyperspectral Aphid Detection via FastGAN Synthetic Data Generation, Transformer-Based Classification and Explainable AI

Ali Saeidan

Lancaster Medical School, Faculty of Health and Medicine, Lancaster University, Lancaster LA1 4YG, United Kingdom

## Abstract

Early detection of aphid infestation in crops is essential for preventing yield loss and reducing unnecessary pesticide use. Hyperspectral imaging combined with Spectral Information Divergence (SID) analysis offers a non-destructive approach for monitoring plant health; however, deep learning methods applied to hyperspectral data are often limited by small dataset sizes. In this study, a data-efficient generative adversarial network (FastGAN) was employed to augment a hyperspectral SID dataset of faba bean leaves containing healthy and aphid-infested samples. The trained generator produced 10,000 synthetic images preserving structural and spectral characteristics of real samples. Image quality was evaluated using Fréchet Inception Distance (FID), demonstrating stable convergence and realistic reconstruction of leaf morphology and infestation patterns. The augmented dataset was used to train four classification architectures: VGG16, ResNet-50, EfficientNet, and Vision Transformer (ViT). Results showed that dataset augmentation significantly improved classification robustness, with performance progressively increasing from classical convolutional networks to transformer-based models. The ViT model achieved the highest accuracy and F1-scores, while EfficientNet provided strong balanced performance and ResNet-50 showed moderate improvements over VGG16. Confusion matrix analysis confirmed reduced false negatives and improved disease detection when using advanced architectures. The findings demonstrate that FastGAN-based augmentation effectively enhances hyperspectral plant disease classification and that transformer-based models provide the most reliable discrimination between healthy and infested leaves.

## 1. Introduction:

Crop protection remains a critical challenge in global food production, as insect pests such as aphids (Aphis fabae) can significantly reduce crop yields and quality. Conventional detection methods rely on visual inspection or chemical analyses, which are labor-intensive, time-consuming, and often destructive (Venkateswara & Padmanabhan, 2025). Hyperspectral imaging (HSI) has emerged as a promising non-invasive tool for monitoring plant health, providing spectral information that captures subtle physiological changes in crops (Karukayil et al., 2026; García-Vera et al., 2024). Spectral Image Decomposition (SID) techniques, in particular, have been applied to identify signatures associated with biotic stresses such as pest infestations. In our previous work (Saeidan et al.,2025). Detection of aphid infestation on faba bean (Vicia faba L.) by hyperspectral imaging and spectral information divergence methods. we demonstrated that SID-based hyperspectral analysis could distinguish between healthy and infested faba bean plants with reasonable accuracy. However, one of the persistent challenges in applying deep learning to HSI data is the limited size of available datasets (Ma et al., 2019; Li et al., 2023). Deep convolutional neural networks (CNNs) and vision transformers (ViTs) require large volumes of labeled data to achieve generalizable performance (Dosovitskiy et al., 2021; Rawat, W., & Wang, Z, 2017). Small datasets not only restrict the complexity of models that can be trained but also increase the risk of overfitting.

Generative Adversarial Networks (GANs) have emerged as a potential solution to this limitation by synthesizing realistic artificial samples that expand the training space (De Rezende et al., 2018). In this work, we employed FastGAN, an efficient and lightweight generative framework specifically designed for high-resolution image synthesis from limited data (Liu et al., 2020). Unlike traditional GAN architectures such as DCGAN or StyleGAN (Karras et al., 2020), FastGAN introduces skip-layer channel-wise excitation (SLE) modules that enable stable training and rich feature propagation even with small datasets. It also uses a simplified discriminator design that reduces computational costs while preserving fidelity. These characteristics make FastGAN particularly suitable for generating synthetic hyperspectral SID images, where the availability of labeled data is limited but high spectral detail is essential. In our study, FastGAN was trained on the real SID dataset to produce 10,000 synthetic images that preserved the spatial and spectral structure of real samples.

The augmented dataset was then used to train and evaluate several deep learning classifiers, each representing a distinct family of architectures. VGG16 (Kulkarni et al., 2024) is a classical deep CNN with 16 weight layers, characterized by small 3×3 convolutional filters and a uniform architecture that has demonstrated strong performance on various vision tasks. ResNet (Islam et al., 2023) introduces residual connections that enable very deep networks to converge by mitigating the vanishing gradient problem, allowing the model to learn complex hierarchical features. EfficientNet (Sangar & Rajasekar, 2025) is a more recent CNN that scales network width, depth, and resolution using a compound scaling coefficient, achieving state-of-the-art accuracy with far fewer parameters than traditional CNNs. Finally, the Vision Transformer (ViT) (Chen et al., 2023; Wang et al., 2025) replaces convolutional operations with a self-attention mechanism that processes images as sequences of non-overlapping patches, enabling the model to capture long-range spatial dependencies and global context.

By comparing these architectures, we aim to assess both classical convolutional and transformer-based approaches under real and GAN-augmented SID datasets. Our central objective is to determine whether FastGAN-generated synthetic images can improve classification accuracy and model robustness in the context of hyperspectral aphid detection.

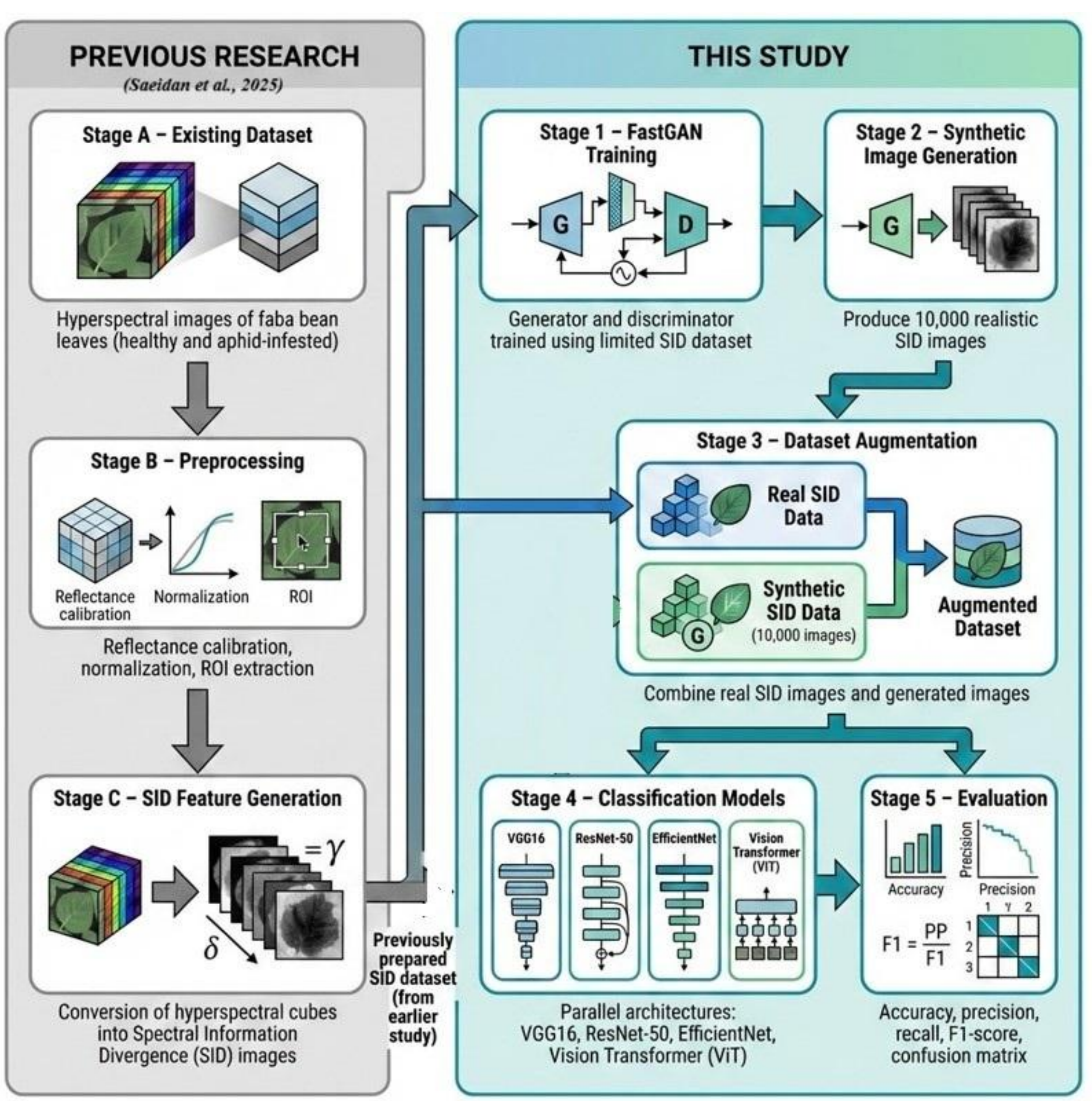

**Figure 1.** Schematic of system.

The main contributions of this study are:

(1) Development of a FastGAN-based framework for generating synthetic SID images from limited hyperspectral datasets of aphid-infested faba bean leaves.

(2) Comprehensive comparison of four representative deep-learning architectures (VGG16, ResNet-50, EfficientNet, and Vision Transformer) using both real and GAN-augmented datasets.

(3) Investigation of the effectiveness of synthetic data augmentation for improving aphid detection robustness under few-shot conditions.

(4) Integration of explainable AI techniques to identify image regions contributing to aphid infestation classification.

## 2. Materials and Methods:

### 2.1 Dataset Preparation

The data set used in this study consisted of hyperspectral Spectral Information Divergence (SID) images of Vicia faba (faba bean) leaves collected under controlled greenhouse conditions. Both healthy and aphid-infested plants were imaged using a hyperspectral camera configured for the visible–near infrared (VIS–NIR) range. Image acquisition and reflectance calibration (white/dark reference correction), region-of-interest (ROI) extraction, and pre-processing followed the protocol established in our previous work, including standard normal variate (SNV) normalization and noise removal. Each hyperspectral cube was spectrally normalized and spatially registered before feature extraction. To ensure comparability, all instrument specifications and acquisition parameters matched those reported earlier, including exposure control, wavelength calibration, and spectral resolution. The dataset was partitioned at the plant level to prevent data leakage, with 70% of samples used for training, 15% for validation, and 15% for testing. The test subset was kept entirely real (non-augmented) to objectively evaluate model generalization. For adapting hyperspectral data to 2D CNN or transformer architectures, two strategies were used: (i) generating three-channel representations using principal component analysis (PCA-RGB) or selected VIS/NIR bands; and (ii) maintaining a narrowband stack focused on wavelengths between 710–825 nm, previously identified as discriminative for aphid stress detection.

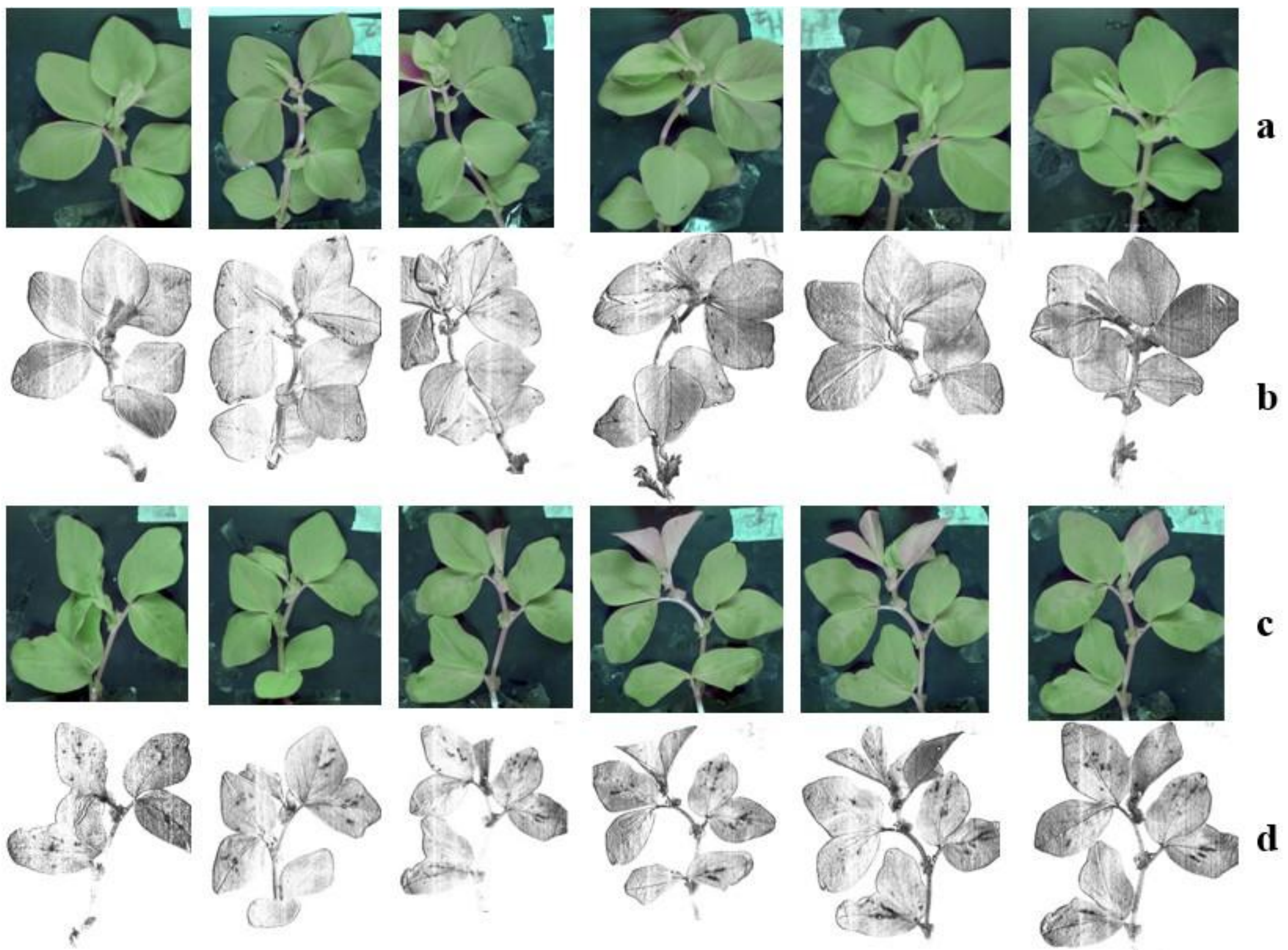

**Figure.2** Some of Original data sets of Healthy RGB images (a) and their corresponding SID images (b), Infected RGB images (c) and their corresponding SID images (d).

## 2.2 Synthetic Data Generation Using FastGAN

To address the limited sample size of the SID dataset, we employed FastGAN, a data-efficient generative adversarial network capable of producing high-quality images under few-shot conditions. Unlike conventional GANs that require large datasets, FastGAN incorporates Skip-Layer Channel-wise Excitation (SLE) modules that enhance gradient propagation and preserve multi-scale contextual information. This architecture stabilizes training and improves fine structural detail reproduction in low-data regimes, making it well-suited for hyperspectral imaging scenarios. The FastGAN model consisted of a generator with hierarchical up sampling and residual convolutional blocks linked through SLE connections, and a discriminator employing multi-scale convolutions with spectral normalization for improved stability. Both networks were trained using the non-saturating logistic loss with an R1 gradient penalty applied to the discriminator every 16 iterations. The Adam optimizer was used for both generator and discriminator with a learning rate of $2\times10^{-4}$, $\beta_1=0.5$, and $\beta_2=0.999$. Mini-batch size was set to 16, and mixed-precision training was used to optimize GPU utilization. To regularize training under data scarcity, we applied differentiable augmentations—including translation, cutout, and brightness/contrast perturbations—to both real and synthetic samples. Additionally, a small replay buffer of generated samples was maintained to prevent catastrophic forgetting. The model was trained for 200–400 epochs, with early stopping based on the validation loss trend. From the converged generator, 10,000 synthetic SID images were generated and visually screened to remove artifacts or unrealistic patterns before inclusion in the classification dataset.

## 2.3 Image Quality Evaluation (FID)

The quality and diversity of the generated images were quantitatively assessed using the Fréchet Inception Distance (FID) metric. FID measures the distance between real and synthetic image feature distributions, defined as: $FID = ||\mu_r - \mu_g||^2 + Tr(\Sigma_r + \Sigma_g - 2(\Sigma_r\Sigma_g)^{1/2})$, where $(\mu_r, \Sigma_r)$ and $(\mu_g, \Sigma_g)$ denote the means and covariances of features extracted from real and generated samples, respectively (Heusel et al., 2017). Because hyperspectral data are not naturally RGB, we used two FID variants: (1) HSI-Encoder FID (primary), in which feature embeddings were derived from a self-supervised spectral encoder (a shallow 3D-CNN trained on real SID cubes); and (2) PCA-RGB FID (secondary), where hyperspectral cubes were reduced to three principal components and passed through an Inception-V3 model. Each FID computation used at least 5,000 generated and 5,000 real samples to ensure statistical robustness.

### 2.4 Classification Models

Four deep learning architectures were used to classify both real and GAN-augmented datasets: VGG16, ResNet-50, EfficientNet-B0/B2, and Vision Transformer (ViT-B/16). All models were initialized with ImageNet-pretrained weights and modified for binary classification (healthy vs. infested). VGG16 employed a stack of 3×3 convolutional layers with ReLU activations, max-pooling, and a two-layer fully connected head with dropout (0.3–0.5). ResNet-50 incorporated residual skip connections to enable deeper gradient flow and reduce vanishing gradients; training used label smoothing ($\varepsilon$=0.1) and cosine learning rate scheduling. EfficientNet-B0/B2 applied compound scaling across network depth, width, and resolution to balance accuracy and efficiency, trained with stochastic depth and data augmentations (RandAugment, mixup=0.2, cutmix=0.2). ViT-B/16 divided each image into 16×16 patches, linearly projected them into tokens, and applied multi-head self-attention to model long-range dependencies. For hyperspectral compatibility, CNNs were provided either PCA-RGB or selected band combinations, while ViT used spectral tokenization treating narrowband groups as token channels. Training used SGD with momentum (0.9) for CNNs and AdamW for ViT, with initial learning rates of 0.01 and $3\times10^{-4}$, respectively, and weight decay of $1\times10^{-4}$. All models were trained under two regimes: Real-Only and Real+GAN, with early stopping based on validation AUC. Evaluation metrics included accuracy, precision, recall, F1-score, and ROC-AUC, alongside calibration error (ECE) and confusion matrices.

### 2.5 Experimental Implementation

All experiments were implemented in PyTorch 2.0 and executed on an NVIDIA GPU cluster (RTX A6000, 48 GB VRAM). Data loading and augmentation were handled using the Albumentations library. Reproducibility was ensured through fixed random seeds and deterministic cuDNN configurations. Training scripts, model checkpoints, and evaluation pipelines were developed for full reproducibility and will be made publicly available upon publication.

### 2.6 Evaluation of results

#### 2.6.1 Explainable AI (Grad-Cam)

Grad-CAM is a gradient-based visualization technique that provides class-discriminative localization maps by leveraging the gradients of a target class flowing into convolutional feature maps. By computing the importance weights through global average pooling of gradients, Grad-CAM highlights regions in the input image that contribute most to the prediction. It is widely applicable across different CNN architectures without requiring architectural modifications. However, due to its reliance on high-level feature maps and spatial averaging, Grad-CAM typically produces coarse and low-resolution heatmaps, which may limit its ability to capture fine-grained or small-scale features(Selvaraju, Cogswell et al. 2017). LayerCAM is an advanced class activation mapping method designed to generate fine-grained visual explanations by leveraging hierarchical feature representations. Unlike Grad-CAM, which computes importance weights through global average pooling of gradients, LayerCAM assigns importance at each spatial location by combining activations with their corresponding gradients in a pixel-wise manner. This allows LayerCAM to preserve detailed spatial information and produce higher-resolution localization maps. Furthermore, by aggregating responses from multiple convolutional layers, LayerCAM captures both high-level semantic information and low-level fine details, making it particularly effective for detecting small or subtle features. Previous studies have demonstrated that LayerCAM significantly improves localization accuracy compared to Grad-CAM, especially in tasks involving fine structures and texture patterns(Jiang, Zhang et al. 2021).

## 3.Results and discussion:

Figure 3 reports the Fréchet Inception Distance (FID) across training iterations for the two leaf conditions, where lower FID indicates that generated images more closely match the real-image distribution. Both models start with high FID (≈510 for healthy, ≈535 for infested), confirming that early-stage generations are far from the target distributions. A clear difference emerges in convergence speed and stability. The healthy-leaf model improves rapidly, dropping below ≈100 FID by ~70–80 iterations and then entering a stable plateau (roughly ≈75–90) through iteration 200, with only minor fluctuations. This indicates earlier and more stable generative convergence for healthy leaves.

In contrast, the infested-leaf model shows strong oscillations for a longer period. After an initial decrease (to ≈185 by ~30 iterations), the curve exhibits repeated spikes (often ≈250–460) between ~40 and ~110 iterations, suggesting training instability and/or greater difficulty modeling the infested-leaf distribution. Only after ~120 iterations does the infested model settle into a low-FID regime comparable to the healthy case (≈85–110), although it remains slightly more variable and ends around ≈100 FID at iteration 200.

Overall, Figure 3 demonstrates that healthy leaves reach low FID earlier and with smoother convergence, while infested leaves require more iterations to stabilize, consistent with the infested domain being more heterogeneous (e.g., variable lesion patterns, discoloration, and texture changes) and therefore harder for the generator to model reliably.

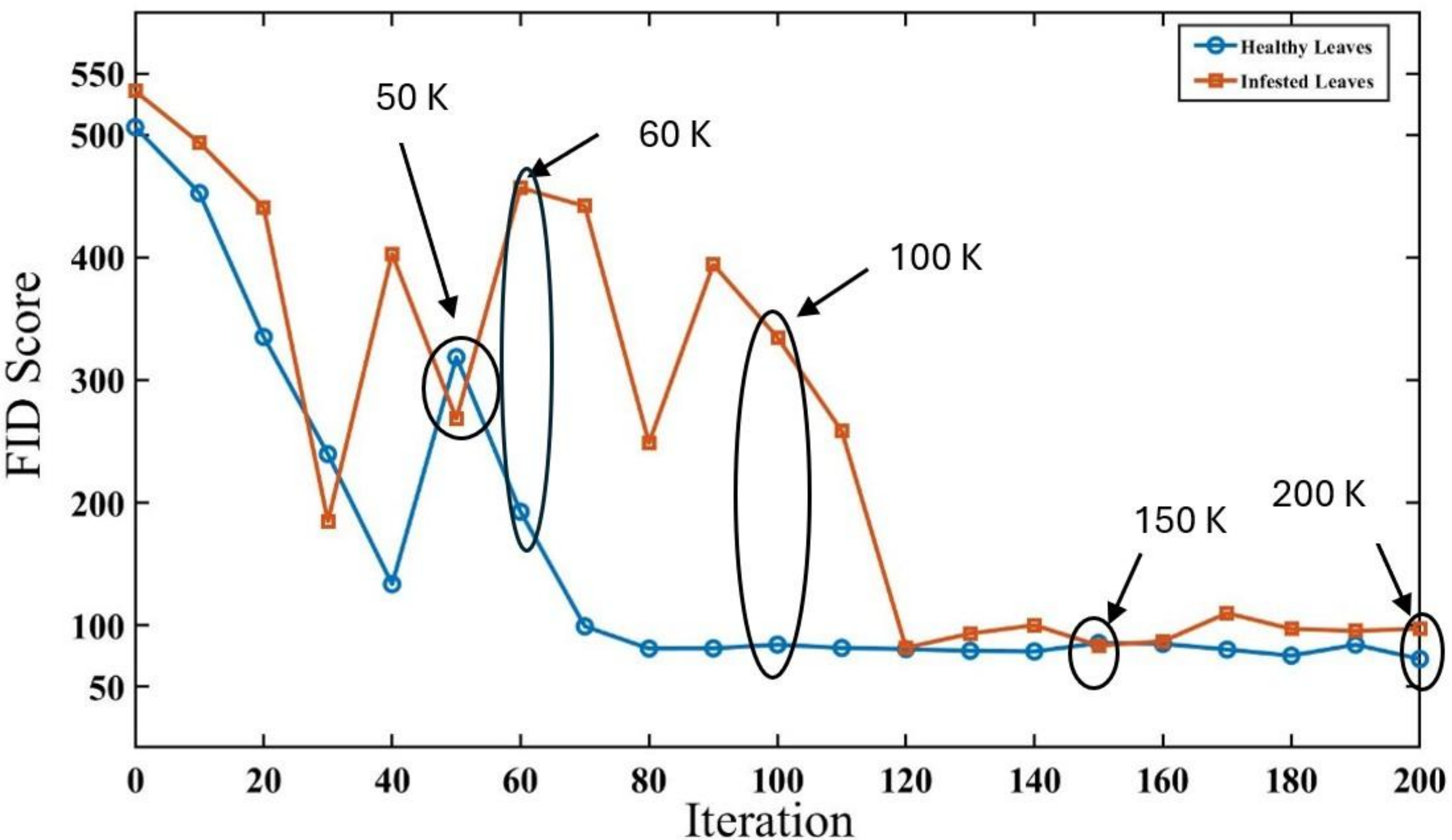

**Figure 3.** FID scores for healthy and infested leaf images as training progresses. Health images reach lower FID values earlier, indicating faster generative model convergence.

Figure 3 qualitatively illustrates how FastGAN outputs evolve from initialization to 200k iterations for healthy and infested leaf domains, alongside representative real samples. At iteration 0, both models generate unstructured, noisy patterns with very high FID (healthy: 506.05; infested:

535.49), indicating minimal similarity to the target distributions. By 50k iterations, coarse leaf silhouettes begin to emerge and background artifacts reduce, reflected by lower FID values (healthy: 318.68; infested: 268.42).

As training continues, the healthy-leaf generator improves more consistently: at 60k the leaf outline and venation become clearer (FID: 192.65), and by 100k the samples closely resemble real leaves with well-defined shape and texture (FID: 84.44). After this point, improvements are mainly refinements, with FID staying low and stable (150k: 85.73; 200k: 72.70), suggesting convergence and visually realistic structure.

In contrast, the infested-leaf generator exhibits a less stable trajectory. Despite some early improvement at 50k, the model shows noticeable artifacts and structural distortions at 60k and 100k, which aligns with the temporary FID increase (60k: 456.65; 100k: 334.73). The outputs become substantially more realistic only after extended training: by 150k, lesion-like texture and leaf morphology are better captured (FID: 83.66). At 200k, the generator maintains realistic overall structure but with slightly higher residual discrepancy (FID: 97.26), consistent with the greater visual variability and complexity of infestation patterns compared to healthy leaves.

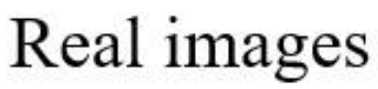

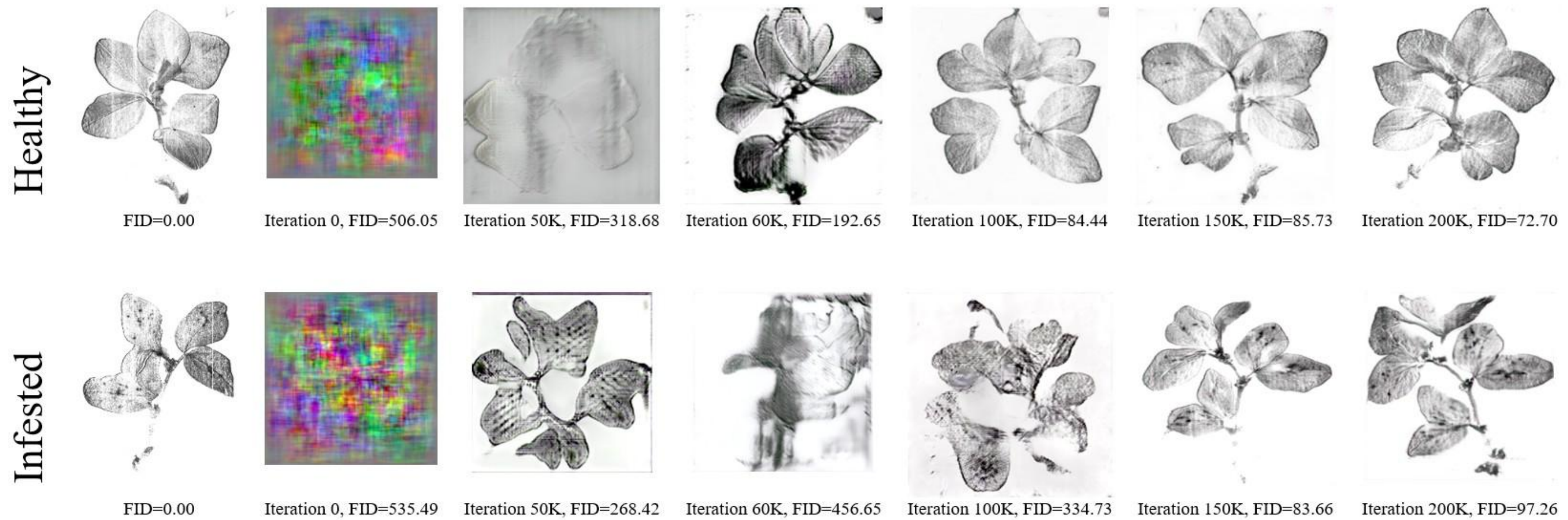

**Figure 3.** Visual progression of FastGAN training from initialization to 200,000 iterations. Real leaf images (FID = 0) are shown in the first column for reference. Generated images at Iterations 0, 50k, 60k, 100k, 150k, and 200k demonstrate the gradual refinement of leaf shape, texture, and venation patterns. The corresponding Fréchet Inception Distance (FID) values decrease consistently, indicating improved similarity between synthetic and real images as the training stabilizes.

Table 1 summarizes the performance of four deep learning architectures (VGG16, ResNet-50, EfficientNet, and Vision Transformer (ViT)) evaluated for healthy vs. infested leaf classification using different training dataset sizes. Performance was measured using precision, recall, F1-score, and overall accuracy.

Across all models, increasing the number of training samples improved classification stability and reduced class imbalance. Very small datasets (e.g., 200 samples) often produced extremely high training accuracy but less consistent generalization, whereas larger datasets provided a better balance between precision and recall. This indicates that greater data variability helps the models learn representative disease patterns and improves robustness.

**Table 1.** Classification results.

| Classifiers | Samples quantity | Train accuracy | Validation accuracy | | Precision | Test Recall | F1-Score | Accuracy |
|---|---|---|---|---|---|---|---|---|
| VGG16 | 200 | 1.00 | 1.00 | Healthy | 0.95 | 1.00 | 0.98 | 0.86 |
| | | | | Infested | 1.00 | 0.95 | 0.97 | |
| | 2000 | 1.00 | 0.99 | Healthy | 0.89 | 1.00 | 0.94 | 0.94 |
| | | | | Infested | 1.00 | 0.88 | 0.93 | |
| | 10000 | 0.73 | 0.72 | Healthy | 0.77 | 1.00 | 0.86 | 0.86 |
| | | | | Infested | 1.00 | 0.66 | 0.78 | |
| | 20000 | 0.99 | 0.99 | Healthy | 0.71 | 1.00 | 0.83 | 0.79 |
| | | | | Infested | 1.00 | 0.58 | 0.74 | |
| ResNet 50 | 200 | 0.98 | 0.97 | Healthy | 0.76 | 0.97 | 0.85 | 0.83 |
| | | | | Infested | 0.96 | 0.69 | 0.80 | |
| | 2000 | 0.99 | 0.99 | Healthy | 0.86 | 1.00 | 0.92 | 0.90 |
| | | | | Infested | 1.00 | 0.83 | 0.90 | |
| | 10000 | 0.99 | 0.99 | Healthy | 0.77 | 1.00 | 0.87 | 0.85 |
| | | | | Infested | 1.00 | 0.71 | 0.83 | |
| | 20000 | 0.99 | 0.99 | Healthy | 0.86 | 1.00 | 0.92 | 0.92 |
| | | | | Infested | 1.00 | 0.83 | 0.91 | |
| Efficient Net | 200 | 0.80 | 0.90 | Healthy | 0.95 | 0.87 | 0.91 | 0.91 |
| | | | | Infested | 0.88 | 0.95 | 0.92 | |
| | 2000 | 0.92 | 0.98 | Healthy | 0.88 | 0.95 | 0.92 | 0.91 |
| | | | | Infested | 0.95 | 0.87 | 0.91 | |
| | 10000 | 0.93 | 0.99 | Healthy | 0.92 | 0.95 | 0.93 | 0.93 |
| | | | | Infested | 0.95 | 0.91 | 0.93 | |
| | 20000 | 0.93 | 0.99 | Healthy | 0.95 | 0.95 | 0.95 | 0.95 |
| | | | | Infested | 0.95 | 0.95 | 0.95 | |
| ViT | 200 | 0.89 | 0.93 | Healthy | 0.80 | 1.00 | 0.88 | 0.87 |
| | | | | Infested | 0.90 | 0.87 | 0.87 | |
| | 2000 | 1.00 | 1.00 | Healthy | 1.00 | 1.00 | 1.00 | 1.00 |
| | | | | Infested | 1.00 | 1.00 | 1.00 | |
| | 10000 | 1.00 | 1.00 | Healthy | 1.00 | 1.00 | 1.00 | 1.00 |
| | | | | Infested | 1.00 | 1.00 | 1.00 | |
| | 20000 | 0.99 | 1.00 | Healthy | 1.00 | 1.00 | 0.94 | 0.93 |
| | | | | Infested | 1.00 | 0.87 | 0.93 | |
| | 100000 | 0.99 | 0.99 | Healthy | 0.96 | 1.00 | 0.97 | 0.97 |
| | | | | Infested | 1.00 | 0.95 | 0.97 | |

VGG16 achieved perfect training accuracy on small datasets (200–2000 samples), but its performance degraded as dataset complexity increased. With 10,000 samples, the F1-scores decreased to 0.86 for healthy leaves and 0.78 for infested leaves, and overall accuracy dropped to 0.86. The model also exhibited lower recall for the infested class at larger sample sizes, indicating difficulty capturing heterogeneous infection patterns. These results suggest overfitting on small datasets and limited adaptability to more diverse data.

ResNet-50 demonstrated more stable behavior than VGG16 across dataset scales. F1-scores remained between 0.83 and 0.92 with accuracies up to 0.92. The residual connections improved generalization and produced a more balanced detection of healthy and infested leaves. However, performance gains plateaued at higher sample sizes, indicating moderate representational capacity compared to more advanced architectures.

EfficientNet consistently achieved strong and balanced results across all dataset sizes. With 20,000 samples, both classes reached an F1-score of 0.95 and overall accuracy of 0.95. The model maintained higher recall for infested leaves than previous CNNs, indicating improved sensitivity to disease characteristics and better handling of intra-class variability.

The ViT model achieved the best performance overall. With 2,000–10,000 samples it reached perfect classification (accuracy = 1.00, F1-score = 1.00), and even with the largest dataset (100,000 samples) it maintained high performance (accuracy = 0.97, F1-score = 0.97). The attention mechanism enabled the model to effectively capture complex spatial patterns associated with leaf infection symptoms, resulting in superior generalization.

Overall, model performance improved progressively from VGG16 to ResNet-50 and EfficientNet, with the Vision Transformer (ViT) consistently delivering the highest accuracy and F1-scores. This indicates that architectural capacity plays a crucial role in plant disease recognition beyond dataset size alone, with transformer-based models providing the most robust classification for healthy versus infested leaf discrimination.

Figure 4 illustrates the variation in classification performance of VGG16, ResNet-50, EfficientNet, and ViT as the number of training samples increases (200, 2K, 10K, and 20K). The figure reports overall accuracy (a), precision for the healthy class (b), and precision for the infested class (c).

Accuracy generally improves with larger datasets, although the magnitude of improvement depends on the architecture. The ViT model achieves the highest performance across nearly all dataset sizes, reaching perfect accuracy at 2K and 10K samples and maintaining high performance at 20K. EfficientNet shows steady improvement with increasing data, achieving the second-best accuracy at larger sample sizes. ResNet-50 demonstrates moderate performance with limited improvement beyond 2K samples. In contrast, VGG16 performance declines after 2K samples, suggesting sensitivity to increased dataset variability and weaker generalization capability.

For healthy leaves, ViT maintains consistently high precision across all dataset sizes and reaches perfect precision from 2K samples onward. EfficientNet provides stable performance with gradual improvement as the dataset grows. ResNet-50 benefits significantly from larger datasets, achieving its best performance at 10K samples before slightly decreasing at 20K. VGG16 shows decreasing precision as the dataset size increases, indicating increased false positives when distinguishing healthy leaves from infested ones.

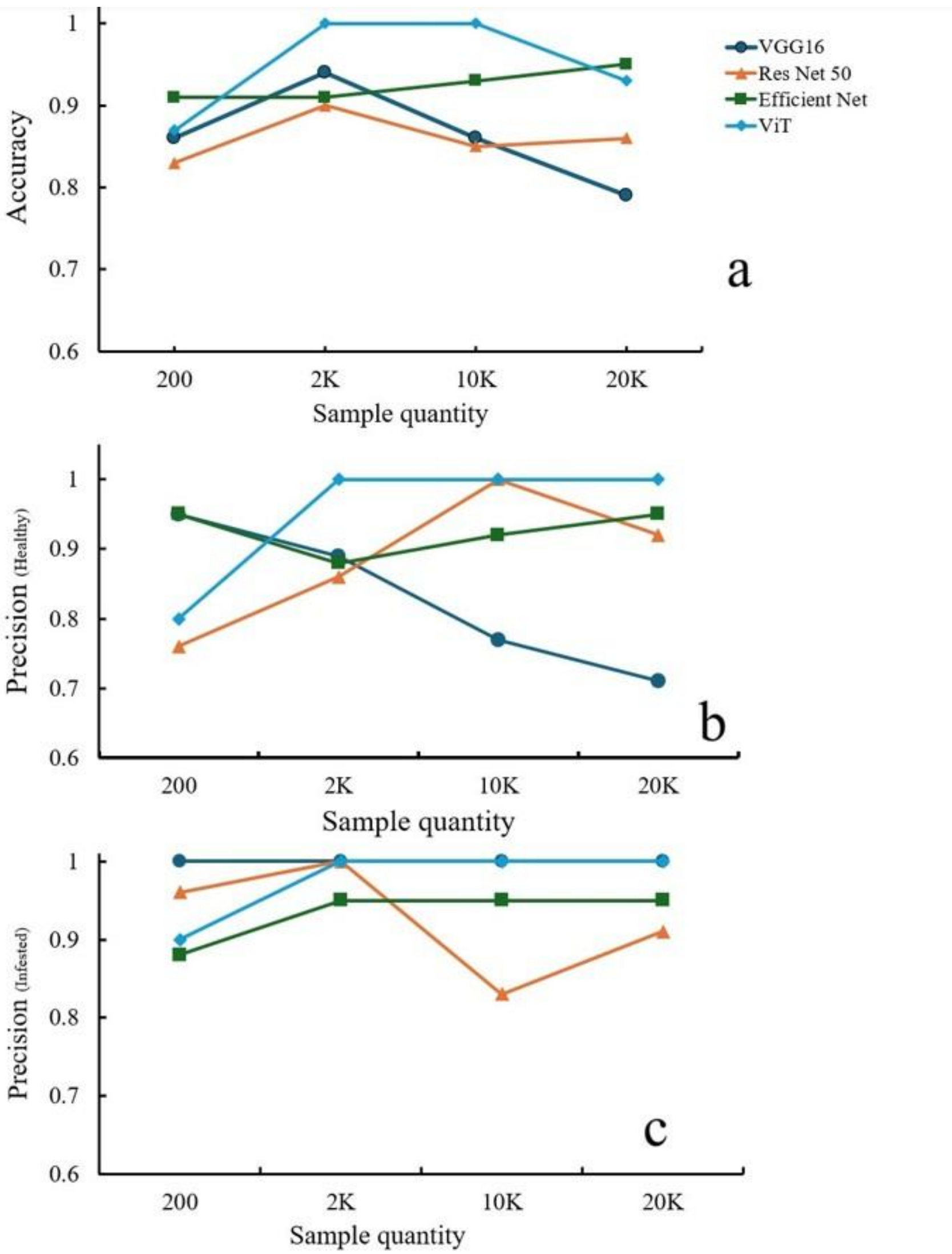

**Figure 4.** The classification performance.

All models perform well on small datasets; however, differences become apparent with larger datasets. ViT consistently achieves perfect or near-perfect precision across all sample sizes, demonstrating strong capability in identifying disease patterns. EfficientNet maintains high and stable precision across dataset scales. ResNet-50 shows fluctuating performance, with a notable drop at 10K samples before recovering at 20K. VGG16 maintains high precision but without improvement, suggesting limited adaptability to increased sample diversity.

Across all metrics, performance improves progressively from VGG16 to ResNet-50 and EfficientNet, with ViT consistently providing the most stable and accurate results. The results indicate that transformer-based architectures better capture complex visual patterns in leaf disease datasets and scale more effectively with increased training data, while traditional CNNs show reduced robustness when dataset variability increases.

Figure 5 shows the confusion matrices for the four evaluated models on the healthy versus infested leaf classification task.

The first model correctly classified all healthy samples (24/24) but struggled to detect infected leaves. Only 14 out of 24 infested samples were correctly identified, while 10 were misclassified as healthy. This indicates strong recognition of healthy tissue but a high false-negative rate for disease detection.

The second model demonstrated improved performance, correctly identifying all healthy leaves and 20 of 24 infested leaves, with only four infected samples predicted as healthy. The results show a better balance between sensitivity and specificity compared to the previous model.

The third model achieved near-perfect classification, correctly detecting 23 healthy and 23 infested samples, with only one misclassification in each class. This indicates a well-balanced capability to distinguish between healthy and diseased leaves.

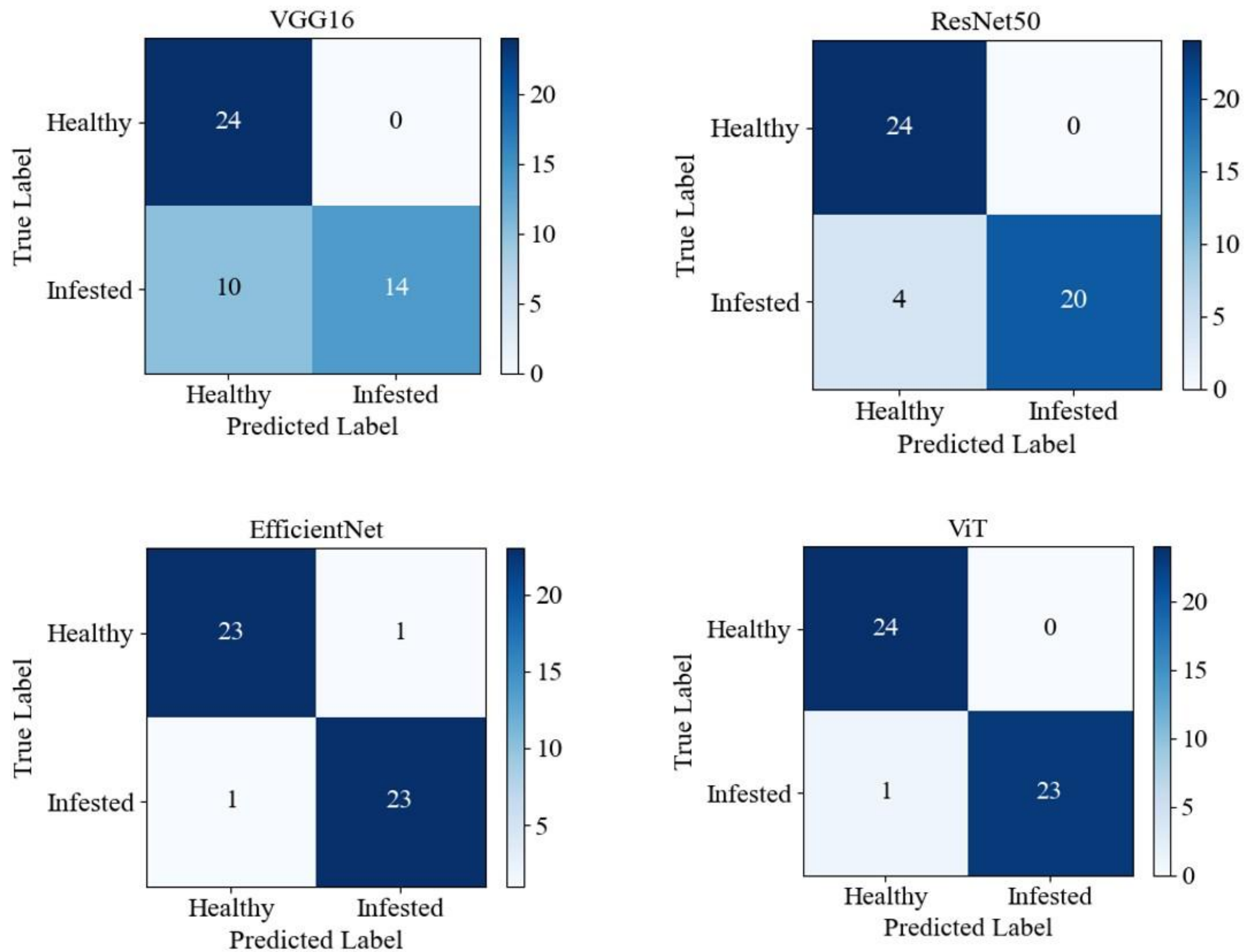

**Figure 5.** Confusion matrixes.

The fourth model produced the most accurate predictions. All healthy samples were correctly classified and only one infected sample was misclassified, resulting in no false positives and minimal false negatives.

Overall, the confusion matrices reveal a progressive improvement in classification reliability across the architectures, with later models reducing both missed infections and incorrect disease predictions, demonstrating superior ability to capture visual disease patterns.

Explainable AI (Grad-Cam)

Figure 6 compares LayerCAM visualizations obtained from different depths of the EfficientNet model to assess their ability to localize aphid infestations. The results show that mid-level representations (feature 5) consistently provide the most accurate and spatially precise localization of small infestation regions across all samples. These heatmaps highlight multiple localized areas on the leaves that correspond to the distribution of aphids. In contrast, feature 6 begins to lose fine-grained detail, focusing on fewer and slightly larger regions. Deeper layers (features 7 and 8) increasingly emphasize global plant structures such as stems and overall leaf arrangement, resulting in diffuse and less relevant activations.

This trend demonstrates that shallow-to-mid feature layers retain critical texture-level information necessary for detecting small-scale biological patterns, while deeper layers capture more abstract semantic features. Overall, feature 5 offers the best balance between spatial resolution and semantic relevance for identifying aphid infestations.

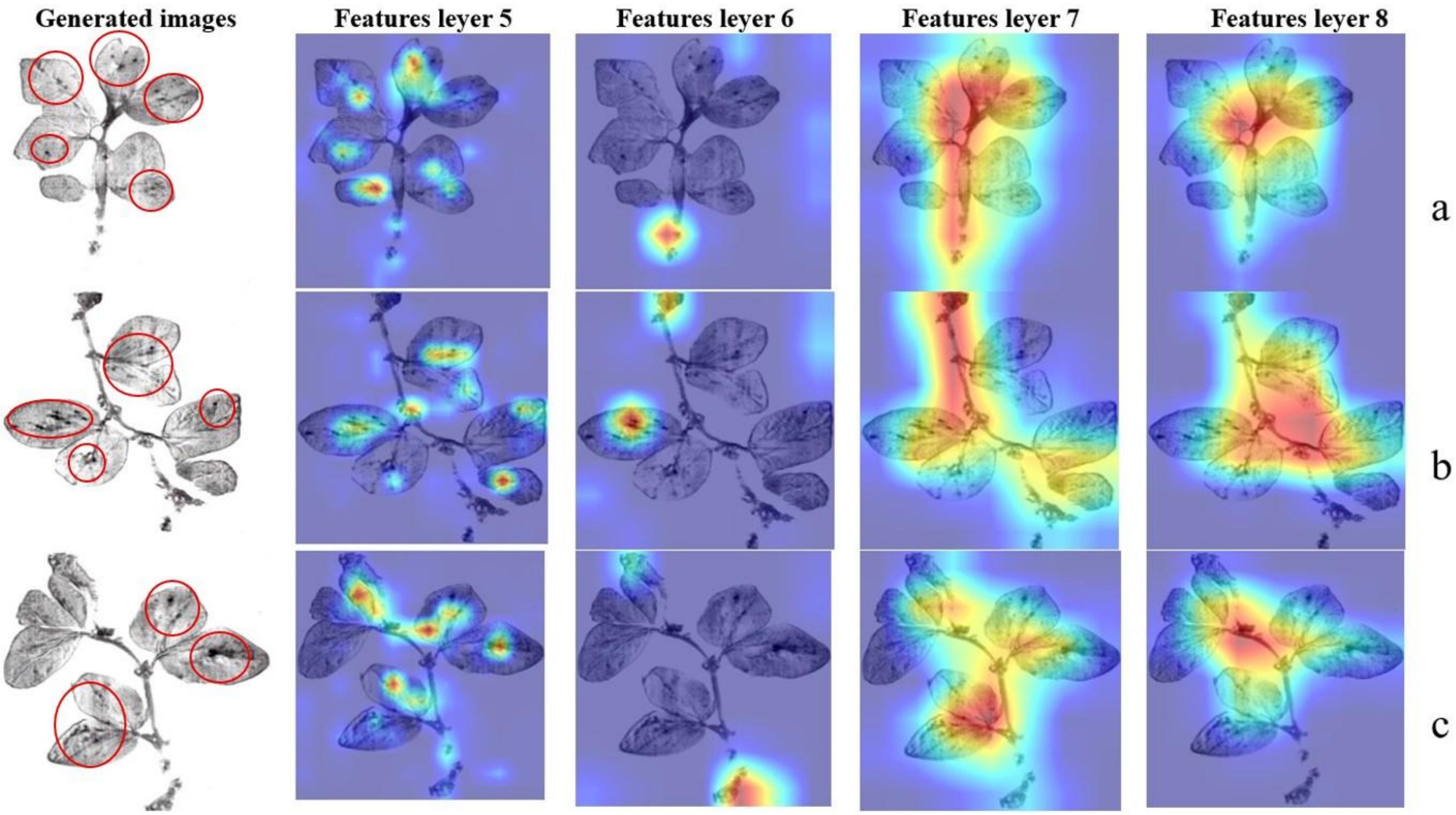

**Figure 6.** Figure 6. LayerCAM visualizations across EfficientNet feature layers for synthetic infested leaf samples. The first column shows input images, while columns 2–5 correspond to feature layers 5, 6, 7, and 8, respectively. Feature 5 provides the most localized detection of infestation patterns, while deeper layers progressively focus on broader plant structures.

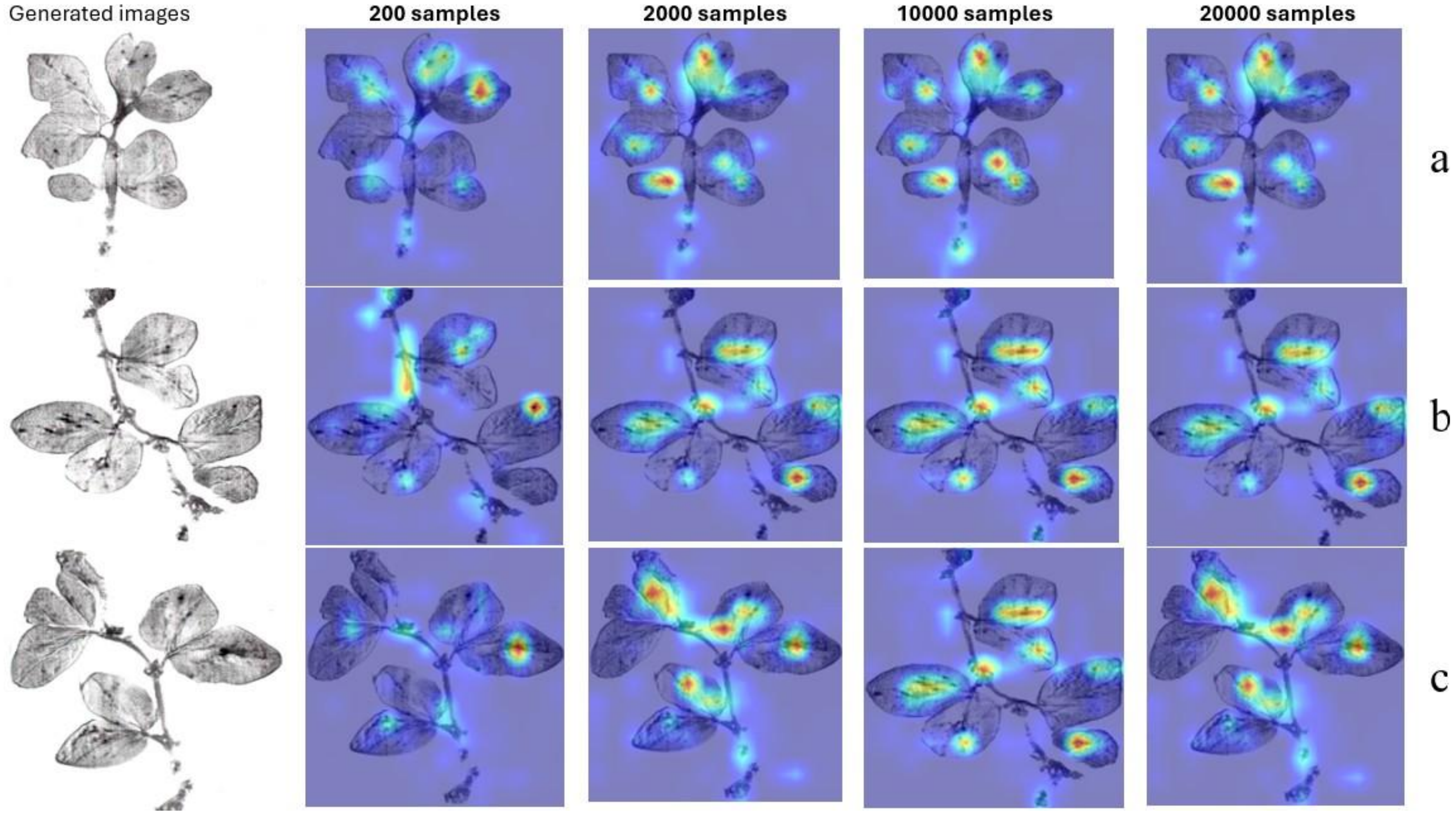

**Figure 7.** EfficientNet model with 200, 2000, 10000 and 20000 samples.

# 4. Discussion

## 4.1 Effectiveness of FastGAN-Based Data Augmentation

One of the major limitations in hyperspectral plant disease detection is the scarcity of labeled training data. Acquiring hyperspectral datasets is often expensive, time-consuming, and labor-intensive, which restricts the development of robust deep learning models. In the present study, FastGAN-generated synthetic SID images improved classification performance across all evaluated architectures, indicating that synthetic data augmentation can effectively address data scarcity in hyperspectral imaging applications.

The observed improvement can be attributed to the increased diversity introduced by the generated samples. By expanding the variability of spectral and structural patterns available during training, FastGAN exposed the classifiers to a broader representation of healthy and aphid-infested leaves. This expanded training space likely reduced model dependence on a limited number of real samples and mitigated overfitting, particularly for architectures that showed instability when trained on smaller datasets. The findings therefore suggest that generative models can provide a practical strategy for enhancing hyperspectral crop-monitoring systems when large annotated datasets are unavailable.

## 4.2 Comparison of Deep Learning Architectures

The comparative analysis revealed a clear performance progression from conventional convolutional neural networks to transformer-based architectures, with the Vision Transformer (ViT) achieving the highest overall classification performance. This result may be explained by the self-attention mechanism employed by transformers, which enables the model to capture long-range spatial dependencies and global contextual information across the entire image.

Such capability is particularly relevant for SID-based hyperspectral representations, where stress responses associated with aphid infestation may be distributed across multiple leaf regions rather than confined to a single localized area. Unlike conventional CNNs, which primarily extract local features through convolutional operations, ViT integrates information from distant image regions simultaneously, allowing more effective recognition of complex spatial patterns. The results therefore suggest that transformer-based approaches may be particularly suitable for hyperspectral plant stress detection tasks that require the integration of distributed spectral and structural information.

EfficientNet also demonstrated strong and stable performance across different dataset sizes. Its compound scaling strategy enables efficient feature extraction while maintaining a relatively low computational burden, making it an attractive alternative for practical deployment. ResNet-50 achieved more balanced performance than VGG16, likely due to the residual connections that facilitate deeper feature learning and improved gradient propagation. In contrast, VGG16 exhibited signs of reduced generalization as dataset complexity increased, suggesting greater sensitivity to overfitting under limited-data conditions.

### 4.3 Challenges Associated with Aphid Infestation Detection

The results indicate that detecting aphid-infested leaves is inherently more challenging than identifying healthy leaves. Healthy samples generally exhibit relatively consistent spectral and structural characteristics, making them easier for classification models to learn and recognize. In contrast, aphid infestation induces a wide range of physiological responses that may vary according to infestation severity, feeding duration, plant condition, and environmental influences.

This variability results in heterogeneous symptom expression, including chlorosis, discoloration, localized tissue damage, and texture alterations. Consequently, the spectral signatures associated with infestation may differ substantially among individual plants, increasing intra-class variability and reducing class separability. Such heterogeneity likely contributed to the greater difficulty observed in modeling infested samples and may explain the slower convergence and increased variability observed during FastGAN training for infected leaves. These findings highlight the importance of robust feature extraction methods capable of capturing subtle and diverse stress responses.

### 4.4 Interpretation of Model Predictions Using Explainable AI

The explainable AI analysis provided valuable insight into the decision-making process of the deep learning models. Grad-CAM successfully identified general regions associated with model predictions; however, the resulting activation maps were relatively coarse and lacked detailed spatial localization. In contrast, LayerCAM generated substantially more refined visual explanations by preserving spatial information from intermediate feature layers.

The activation patterns suggest that intermediate network layers capture localized texture and structural alterations associated with aphid-induced stress, whereas deeper layers focus more strongly on broader morphological characteristics. This observation reflects the hierarchical feature-learning process of deep neural networks, where low- and mid-level features capture fine details and deeper layers encode higher-level semantic information. The improved localization achieved by LayerCAM may therefore enhance the biological interpretability of model predictions and increase practitioner confidence in future precision-agriculture applications.

# 5. Conclusions and future works

## 5.1 Conclusions

This study investigated a data-efficient framework for hyperspectral aphid infestation detection in Vicia faba leaves through the integration of FastGAN-based data augmentation, deep learning classification, and explainable artificial intelligence. The results suggest that synthetic SID image generation can effectively alleviate the limitations associated with small hyperspectral datasets and improve classification robustness across multiple deep learning architectures.

Among the evaluated models, the Vision Transformer achieved the strongest overall performance, indicating the potential advantages of self-attention mechanisms for capturing complex spectral–spatial patterns associated with aphid-induced stress. EfficientNet also demonstrated competitive performance while maintaining computational efficiency. Furthermore, the explainable AI analysis showed that LayerCAM provided more precise and biologically meaningful visual interpretations than Grad-CAM, improving the transparency of model predictions.

Overall, the findings suggest that combining hyperspectral imaging with generative data augmentation and transformer-based learning provides a promising approach for early, non-destructive aphid detection in precision agriculture. Such approaches may contribute to improved crop monitoring, reduced pesticide use, and more sustainable pest-management strategies.

## 5.2 Limitations and Future Research Directions

Although the proposed framework demonstrated promising performance for hyperspectral aphid detection, several limitations should be acknowledged. First, the study was conducted using a single crop species (Vicia faba) and a single pest species (Aphis fabae). Consequently, the generalizability of the proposed approach to other crop–pest systems remains to be fully investigated. Future studies should evaluate the framework across multiple crops, pest species, and environmental conditions to assess its broader applicability.

Second, the generated synthetic images were primarily evaluated using image-distribution similarity metrics and downstream classification performance. While these assessments provide useful indicators of image quality and practical utility, additional spectral validation approaches could provide deeper insight into the preservation of hyperspectral characteristics within the generated samples. Future work may therefore incorporate more comprehensive spectral fidelity assessments to further validate the biological realism of synthetic hyperspectral data.

Third, all experiments were performed under controlled imaging conditions. Although controlled environments reduce experimental variability and facilitate model development, real-world agricultural settings often involve substantial variation in illumination, background complexity, plant growth stage, and environmental stress factors. Consequently, further validation under greenhouse and field conditions is required to determine the robustness and operational reliability of the proposed framework.

Finally, this study focused on binary classification of healthy and aphid-infested leaves. In practical agricultural applications, crops may simultaneously experience multiple biotic and abiotic stresses with varying severity levels. Future research should therefore explore multi-class classification, infestation severity estimation, and the integration of full hyperspectral cube information to enable more comprehensive plant health monitoring systems.

Despite these limitations, the present study demonstrates the potential of combining hyperspectral imaging, generative data augmentation, transformer-based learning, and explainable artificial intelligence for data-efficient pest detection. Future developments incorporating domain adaptation, physics-informed learning, and cross-dataset validation may further improve model robustness and support the deployment of intelligent hyperspectral monitoring systems in precision agriculture.

## References:

[1] S. M. Venkateswara and J. Padmanabhan, "Deep learning-based agricultural pest monitoring and classification," *Scientific Reports*, 2025.

[2] Karukayil, A., Mota, J. F., & Cheein, F. A. (2026). 3D crop reconstruction: A review of hyperspectral and multispectral approaches. *Computers and Electronics in Agriculture*, *241*, 111282.

[3] García-Vera, Y. E., Polochè-Arango, A., Mendivelso-Fajardo, C. A., & Gutiérrez-Bernal, F. J. (2024). Hyperspectral image analysis and machine learning techniques for crop disease detection and identification: A review. Sustainability, 16(14), 6064.

[4] Saeidan, A., Caulfield, J., Vuts, J., Yang, N., & Fisk, I. (2025). Detection of aphid infestation on faba bean (Vicia faba L.) by hyperspectral imaging and spectral information divergence methods. Journal of Plant Diseases and Protection, 132, Article 109.

https://doi.org/10.1007/s41348-025-01100-6

[5] Li, S., Song, W., Fang, L., Chen, Y., Ghamisi, P., & Benediktsson, J. A. (2019). Deep learning for hyperspectral image classification: An overview. IEEE transactions on geoscience and remote sensing, 57(9), 6690-6709.

[6] Ma, L., Liu, Y., Zhang, X., Ye, Y., Yin, G., & Johnson, B. A. (2019). Deep learning in remote sensing applications: A meta-analysis and review. ISPRS journal of photogrammetry and remote sensing, 152, 166-177.

[7] Li, W., Fu, H., Yu, L., & Cracknell, A. (2016). Deep learning based oil palm tree detection and counting for high-resolution remote sensing images. Remote sensing, 9(1), 22.

[8] Dosovitskiy, A., Beyer, L., Kolesnikov, A., Weissenborn, D., Zhai, X., Unterthiner, T., ... & Houlsby, N. (2020). An image is worth 16x16 words: Transformers for image recognition at scale. arXiv preprint arXiv:2010.11929.

[9] Rawat, W., & Wang, Z. (2017). Deep convolutional neural networks for image classification: A comprehensive review. Neural computation, 29(9), 2352-2449.

[10] De Rezende, E. R., Ruppert, G. C., Theophilo, A., Tokuda, E. K., & Carvalho, T. (2018). Exposing computer generated images by using deep convolutional neural networks. Signal Processing: Image Communication, 66, 113-126.

[11] Liu, B., Zhu, Y., Song, K., & Elgammal, A. (2020, October). Towards faster and stabilized gan training for high-fidelity few-shot image synthesis. In International conference on learning representations.

[12] Karras, T., Aittala, M., Hellsten, J., Laine, S., Lehtinen, J., & Aila, T. (2020). Training generative adversarial networks with limited data. Advances in neural information processing systems, 33, 12104-12114.

[13] Kulkarni, P., Sarwe, N., Pingale, A., Sarolkar, Y., Patil, R. R., Shinde, G., & Kaur, G. (2024). Exploring the efficacy of various CNN architectures in diagnosing oral cancer from squamous cell carcinoma. MethodsX, 13, 103034.

[14] Islam, M. M., Adil, M. A. A., Talukder, M. A., Ahamed, M. K. U., Uddin, M. A., Hasan, M. K., ... & Debnath, S. K. (2023). DeepCrop: Deep learning-based crop disease prediction with web application. Journal of Agriculture and Food Research, 14, 100764.

[15] Sangar, G., & Rajasekar, V. (2025). Optimized classification of potato leaf disease using EfficientNet-LITE and KE-SVM in diverse environments. Frontiers in plant science, 16, 1499909.

[16] Chen, W., Ayoub, M., Liao, M., Shi, R., Zhang, M., Su, F., ... & Wong, K. K. (2023). A fusion of VGG-16 and ViT models for improving bone tumor classification in computed tomography. Journal of Bone Oncology, 43, 100508.

[17] Wang, Y., Deng, Y., Zheng, Y., Chattopadhyay, P., & Wang, L. (2025). Vision transformers for image classification: A comparative survey. Technologies, 13(1), 32.

[18] Heusel, M., Ramsauer, H., Unterthiner, T., Nessler, B., & Hochreiter, S. (2017). Gans trained by a two time-scale update rule converge to a local nash equilibrium. Advances in neural information processing systems, 30.

Jiang, P.-T., et al. (2021). "Layercam: Exploring hierarchical class activation maps for localization." IEEE transactions on image processing 30: 5875–5888.

Selvaraju, R. R., et al. (2017). Grad-cam: Visual explanations from deep networks via gradient-based localization. Proceedings of the IEEE international conference on computer vision.